\documentclass[a4paper,conference]{IEEEtran}

\usepackage{balance}

\usepackage{placeins}

\IEEEoverridecommandlockouts 
\usepackage{graphicx}
\usepackage{algorithmic}
\usepackage{algorithm}
\usepackage{listings}
\usepackage[cmex10]{amsmath}
\usepackage{url}
\usepackage{multirow}

\usepackage[utf8]{inputenc}

\usepackage{amsfonts}

\title{Optimisation using Natural Language Processing:\\Personalized Tour Recommendation for Museums}
\author{\IEEEauthorblockN{Mayeul Mathias \IEEEauthorrefmark{1},
Assema Moussa \IEEEauthorrefmark{1},
Fen Zhou \IEEEauthorrefmark{1}\IEEEauthorrefmark{3},
Juan-Manuel Torres-Moreno \IEEEauthorrefmark{1}\IEEEauthorrefmark{3}\IEEEauthorrefmark{4},
\\
Marie-Sylvie Poli \IEEEauthorrefmark{2}\IEEEauthorrefmark{3},
Didier Josselin \IEEEauthorrefmark{1}\IEEEauthorrefmark{3}\IEEEauthorrefmark{5},
Marc El-B\`eze \IEEEauthorrefmark{1}\IEEEauthorrefmark{3},
Andr\'ea Carneiro Linhares \IEEEauthorrefmark{6},
Françoise Rigat \IEEEauthorrefmark{7}}
\IEEEauthorblockA{\{mayeul.mathias, assema.moussa\}@alumni.univ-avignon.fr
\\
\{fen.zhou, juan-manuel.torres, marie-sylvie.poli, didier.josselin, marc.elbeze\}@univ-avignon.fr
\\
andrea.linhares@ufc.br, francoise\_rigat@yahoo.it}
\IEEEauthorblockA{\IEEEauthorrefmark{1}LIA, Université d'Avignon et des Pays de Vaucluse, France.}
\IEEEauthorblockA{\IEEEauthorrefmark{2}CNE, Université d'Avignon et des Pays de Vaucluse, France.}
\IEEEauthorblockA{\IEEEauthorrefmark{3}FR 3621 Agorantic - CNRS Université d'Avignon et des Pays de Vaucluse, France.}
\IEEEauthorblockA{\IEEEauthorrefmark{4}\'Ecole polytechnique de Montréal, (Québec) Canada}
\IEEEauthorblockA{\IEEEauthorrefmark{5}UMR 7300 ESPACE - CNRS, France}
\IEEEauthorblockA{\IEEEauthorrefmark{6}Universidade Federal do Cear\'a, Brazil.}
\IEEEauthorblockA{\IEEEauthorrefmark{7}Università degli Studi di Torino, Italy.}
}

\begin{document}
\maketitle

\begin{abstract}
This paper proposes a new method to provide personalized tour recommendation for museum visits. It combines an optimization of preference criteria of visitors with an automatic extraction of artwork importance from museum information based on Natural Language Processing using textual energy. This project includes researchers from computer and social sciences. Some results are obtained with numerical experiments. They show that our model clearly improves the satisfaction of the visitor who follows the proposed tour. This work foreshadows some interesting outcomes and applications about on-demand personalized visit of museums in a very near future.
\end{abstract}

\section{Introduction}

\IEEEoverridecommandlockouts\IEEEPARstart{M}{useums} are no longer only institutions that acquire, store and expose our heritage. Going to a museum is a learning activity but also an enjoyment for visitors. With the emergence of the Web, curators and cultural mediators decided to get involved in collaborative and numerical culture to attract a larger public. Today, almost all museums have a website but few of them allow the visitors to prepare their visit in the best conditions.

Some art, science and society museums are collaborating with research laboratories to develop new technologies that improve services in museums in response to the desires of existing and potential visitors.

However, there are still difficulties, epistemological barriers, to study the expectations and the intentions of different publics, including online visitors. Knowing why people want to come and visit museums could allow automatic systems to suggests their tour, save their time and give them the best of the knowledge of the exhibited arts.

Among all possibilities, a recommendation system for personalized routing is by far one of the best improvements. Indeed, some museums exhibit thousands of artworks and it is not conceivable for a visitor to admire all of them because he might spend time in front of artworks which do not match his interests and he might not be able to see other more interesting artworks due to tiredness or a lack of time. A few museums, as The Louvre, offer a recommendation system\footnote{\url{http://www.louvre.fr/en/parcours}} but they are limited to the selection of a route in a pre-established set. Moreover, in this particular case, the personalization is restricted to the selection of a theme and the duration of the visit in a set of no more than 10 themes and 4 different durations.

It is essential to propose a personalized route for each visitor or group of visitors according to their interests while taking into account their constraints such as limited schedule, physical handicap or a list of artworks to include of the tour. This operation may also reduce unuseful moves (avoid round trips). But to calculate an optimal tour, we need to assess the visitor interest for each artwork by asking his preferences.

Modeling the  preferences with random distributions may not reflect reality because curators take care of the scenography (therefore the coherence) of each room. So we worked on prefered artists (the visitor can select a set of interesting artists) and we propose to use the artworks description to highlight a kind of intrinsic interest from the point of view of the museum. Indeed, the description displayed to the visitor should show how significant is the artwork for the museum. We valuate each item by analyzing their description (with Automatic Text Summarization) and use it as a base value, considering that even without any preference, some artworks are more interesting than others.

\subsection*{The Musée de l'Orangerie}

Due to the time needed to extract and check all the data we worked on this small museum to test our model.

\begin{figure*}
\centering
\includegraphics[scale=0.5]{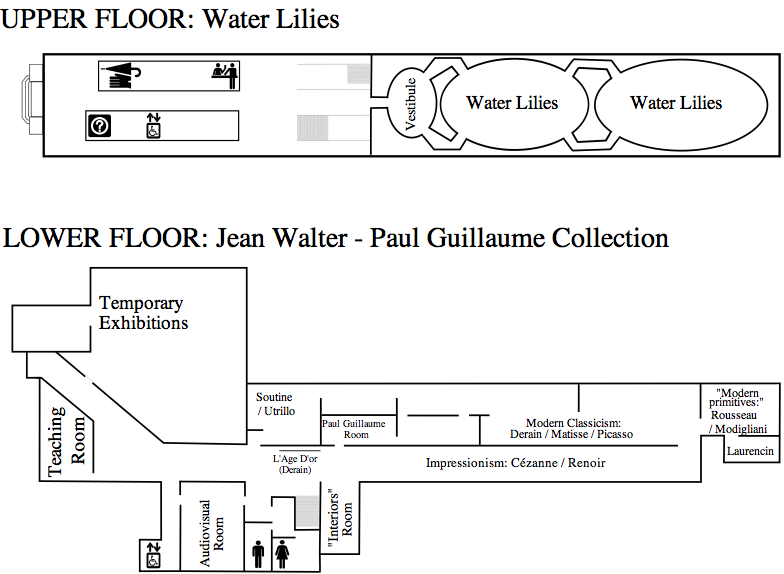}
\caption{Map of the Musée de l'Orangerie}
\label{fig:orangerie-map}
\end{figure*}

The Musée de l'Orangerie (Museum of Orangerie), in Paris (France), regroups 144 artworks from 14 artists in 10 exhibition rooms. The website\footnote{\url{http://musee-orangerie.fr}} of the museum supplies a map (shown in Fig. \ref{fig:orangerie-map}) and indexes information about all the artworks including the name of the artist, a description of the artwork and its date of creation.

\subsection*{Paper organization}

The remaining of this paper is organized as follows. We review the related work in Section \ref{sec:related-work}. In Section \ref{sec:enertex}, we present a Natural Language Processing (NLP) based approach to compute artwork interest. The Personalized Tour Recommendation Problem is presented in Section \ref{sec:personalized-tour-recommendation-problem}. To solve this problem, we develop an Integer Linear Programming based method to solve the tour optimization problem in Section \ref{sec:model} and define a model to represent the visitor preferences in Section \ref{sec:visitor-preferences-modeling}. The simulations are conducted and numerical results are demonstrated in Section \ref{sec:simulations-numerical-results}. Finally, we conclude the paper in Section \ref{sec:conclusion}.

\section{Related work}
\label{sec:related-work}

A first model developed in 2010 \cite{museum_visitor_routing_problem} proposes to formulate the visitor routing problem as an extension of the open shop scheduling problem (in which each visitor group is a job and each interesting room is a machine). Each visitor group has to pass through all rooms but it is impossible for two groups of visitors to be simultaneously in the same room. This restriction can lead to non optimal or infeasible solutions if there are more visitor groups than rooms in the museum (which is the case if we consider each single visitor as a group).

Relying on the constraint programming model \cite{artois_2012}, we propose to reduce the number of used variables. In \cite{artois_2012}, they generate a route by calculating the smallest number $K$ of steps required to cross the museum (to visit all the rooms). This model requires that each artwork is represented as $K$ variables (one per step). Due to the fact that museums often have several thousands of artworks, it leads to a huge number of variables. Moreover they use mathematical distributions to simulate a visitor profile which does not necessary reflect reality (in museums, artworks are often grouped in a room because they are related to each other, a configuration that a random distribution as they used cannot represent).

In 2013, some works \cite{peloponnese_2013} used the visiting style of visitors (the way a visitor go from an artwork to another) but their model requires two matrices of size $N^2$ (where $N$ is the number of artworks). The first one indicates the accessibility to an artwork from another (if they are in the same room or in two rooms directly connected) and the second one contains the distance between two artworks. However as the number of artworks is always greater than the number of rooms, most of the museums are modeled as two sparse matrices with duplicated data (in a room, it is often allowed to freely move between artworks). This makes the use of constraint programming expensive.

\section{Artwork description analysis using Textual Energy}
\label{sec:enertex}

Our idea is to use the description of each artwork as an independent measure of their interest. 
Indeed, two similar artworks (same theme, support, artist) will produce the same result but may be very different. 
By analyzing the description provided by the museum, we tried to differentiate them.

Automatic Text Summarization (ATS) techniques by extraction \cite{luhn:58,sparck:2001,torres:11} allow to rank a set of textual segments (sentences, paragraphs etc.) depending on a measure of similarity. 
Textual Energy algorithm (Enertex) converts a textual document into a physical object and use Statistical Physics to measure its energy  \cite{micai-2007}. 
This energy, to which we should refer as Textual Energy, is then computed and apply to summarization. 
The physical model of Textual Energy gives rise to a single non iterative algorithm of low complexity.
Therefore Textual Energy allows to redefine sentence ranking on simple and efficient matrix operations.
The resulting algorithms are much easier to apply to large texts and give better results without using any post-processing.

\subsection{Starting point: Hopfield Model}
Hopfield's approach \cite{Hopfield1982,Hertz91} was based on magnetic Ising model to build a Neural Network (ANN) with pattern learning capabilities.
The capacities and limitations of this ANN (an associative memory), were well established in a theoretical framework \cite{Hopfield1982,Hertz91}:
the patterns must not be correlated to obtain free error recovery,
the system saturates quickly and only a little fraction of the patterns can be stored correctly.
Despite these major drawbacks, Hopfield contributed to ANN theory by introducing the concept of energy by analogy with magnetic systems.
A magnetic system is a set of $N$ spins like small magnets that can adopt several orientations.
The simplest model is the dipole one or Ising model where there are only two opposite possible orientations:
up ($\uparrow$ or +1) or down ($\downarrow$ or 0).
Ising magnetic model was used in a large variety of systems that can be completely described by a set $N$ of binary variables \cite{Ma1985} with  $2^N$ possible configurations (patterns).
The spins are equivalent to neurons that can interact following Hebb's rule\footnote{Hebb \cite{Hertz91} suggested that synaptic connections change according to the correlation between neuronal states.}:
\begin{equation}
    J_{i,j}= \sum^P_{\mu=1} \left(s_{\mu,i}\times s_{\mu,j}\right)
    \label{eq:J}
\end{equation}
$s_{\mu,i}$ and $s_{\mu,j}$ are the  states of neurons $i$ and $j$ in the pattern $\mu$.
The summation concerns the $P$ patterns to store.
This rule of interaction is local, because $J_{i,j}$ depends only on the states of the connected unities.
It has the capacity to store and to recover certain number of configurations of the system, because the Hebb rule transforms these configurations into attractors (minimal local) of  the energy function \cite{Hopfield1982}:
\begin{equation}
    E_{\mu} = -\frac{1}{2} \sum^N_{i=1}\sum^N_{j=1} \left(s_{i} \times J_{i,j} \times s_{j} \right)
    \label{eq:E}
\end{equation}

The fundamental concept of magnetic energy is a function of the system configuration, that is, of the state of activation or non-activation of its units.
The concept of energy induces a type of interaction.
If we present a pattern $\nu$, every spin will undergo a local field:
$h^i=\sum^N_{j=1} J^{i,j} s^j$
induced by the energy of the others spins.
Therefore the total energy of the new system made of the new pattern inserted into the previous system reflects the interaction  between the pattern and the initial system.

We shall focus on theoretical objects that are usually considered in Statistical Physics.
In magnetic system analysis, these are energy function distributions \cite{molina2013analysis}.
Hopfield himself used these functions to show that the recovery is convergent.
Our Enertex system is entirely grounded on them.

\subsection{Energy as a document similarity measure}

The Vector Space Model (VSM) has also been applied to texts since \cite{salton:83} following a bag of word representation of sentences.
In this model vectors  represent sentences and a document gives rise to a matrix.
We have used VSM to represent documents in our model magnetic system: 
a sentence (a row vector) is equivalent to a Ising spin chain and a document (a magnetic system) is represented by a matrix of $P$ rows $\times$ $N$ columns.
Therefore, sentences can be studied as Ising spin chains.
More formally, with a vocabulary of $N$ words (terms) in a document, it is possible to represent a sentence as a chain of $N$ spins,
$i=1,\cdots,N$.
A document with $P$ sentences is formed of $P$ chains in the vector space $\Xi$ of dimension $N$.
In this paper, the description of each artwork is assimilated as a long pseudo-sentence.
Therefore, a document (the collection of a museum) is constituted of a set of $P$ (pseudo-)sentences.

Documents are preprocessed by removing functional words (by using a stop list), normalized and lemmatized \cite{stem:1,manning99foundations}.
This preprocessing reduces considerably the document dimensionality.
Let be $T=\{t_1,\ldots,t_N\}$ the set of remaining terms after this preprocessing.
Once segmented into units, usually sentences, the text is represented by a set $S=\{\vec{s}_{1},\ldots,\vec{s}_{\mu},\ldots, \vec{s}_P\}$ where each $\vec{s}_\mu$ is the bag of words in segment $\mu$.
As usual in text vector model, we consider the matrix $S_{[P\times N]}=(s_{\mu,j})_{1\leq \mu \leq P, 1\leq j \leq N}$ of  frequency/absence associated to $H$ by:

\begin{eqnarray}
    s_{\mu,j}
    &= &
    \left\{
    \begin{array}{ll} 
        tf_{\mu,j} & \textrm{if } t_j \in  \mbox{ sentence } \mu \\
          0 & \textrm{otherwise}
    \end{array}
    \right.
    \label{taln:matrice:1}
\end{eqnarray}
\noindent where $tf_{\mu,j}$ is the term frequency of $t_j$ in the sentence $\mu$.

We therefore consider the presence of term $t_j$ as a spin $s_j$ $\uparrow$ with magnitude $tf_{\mu,j}$ (its absence by a $\downarrow$ respectively), and a description of each artwork (text segment) by a chain of $N$ spins.

It is common to consider that these vectors are correlated according to the shared words.
Here the introduction of the magnetic model induces moreover indirect interactions.
In this model sentences that do not share any word could however interact because of the magnetic field generated by the other sentences of the document that form the global magnetic system.

We have studied the interactions between the terms and the sentences using Hebb's rule and Ising energy respectively.
To obtain the matrix $J$ of interactions between the $N$ terms, we apply  Hebb's rule (equation \ref{eq:J}) in its matrix form:
\begin{equation}
       J= S^T \times S
    \label{eq:matrice_J}
\end{equation}
where $J_{i,j}$ is the number of co-occurrences of terms in sentences.
The energy function (equation \ref{eq:E}) of a (magnetic) system $S$ is:
\begin{equation}
    E =  S \times J \times S^T = (S \times S^T)^2
    \label{eq:matrice_E}
\end{equation}
Each element  $E_{\mu, \nu}$ represents the energy between sentences  $\mu$ and $\nu$.
The values in the first matrix diagonal quantify the interaction energy between words into a sentence meanwhile the other values in the rest of the matrix show the interactions between distinct sentences.
The sum of absolute values in one row gives the total energy of interaction of the corresponding sentence $\mu$ with the document:
\begin{eqnarray}
    E_{\mu}&=&\sum_\nu \left|e_{\mu,\nu}\right|
\end{eqnarray}

We use this energy value to rank sentences (description of artwork) by order of decreasing importance.
The most energetic will be considered as the most important.

\section{Personalized Tour Recommendation Problem}
\label{sec:personalized-tour-recommendation-problem}

The Personalized Tour Recommendation Problem (PTRP) can be viewed as an optimization problem and solved by optimization techniques. For this purpose, we first model the museum topology as a graph and then formulate the studied problem as an Integer Linear Programming (ILP) instance. Therefore, the optimal personalized tour can be obtained by solving the ILP model we propose.

\subsection{Museum modeling}

A museum is modeled as a 7-tuple $G = \langle V, A, E, X, P, r \rangle$ where:

\begin{itemize}
\item $V$ is the set of vertices, each vertex is an exhibition room, an entrance or an exit of the museum.

\item $A$ is the set of arcs which connect different rooms. There is an arc $a_{ij} \in A$ between two vertices $i$ and $j$, if we can go from room $i$ to room $j$ directly without passing through other rooms.

\item $E$ is the set of entrances of a museum, which is a subset of $V$, i.e. $ E \subset V$.

\item $X$ is  the set of exits of a museum, which is also a subset of $V$, i.e. $ X \subset V$.

\item $P$ is the set of all artworks in the museum.

\item $r$ is a mapping function $P \rightarrow V$. For each artwork $p \in P$, $r(p)$ is the room containing $p$.

\end{itemize}

Some large museums may have several entrances and exits, that is why $E$ and $X$ are two subsets of $V$. We also admit that there is always a path from any entrance to an exit.
We consider directed arcs and not edges because some museums may impose a flow direction for several reasons (minimizing congestion, pedagogical tour). Note that by definition of $A$, there is no incoming arc to any entrance and there is no any outgoing arc from any exit neither.

\subsubsection*{Application to the Musée de l'Orangerie}

\begin{figure}
\centering
\includegraphics[scale=0.75]{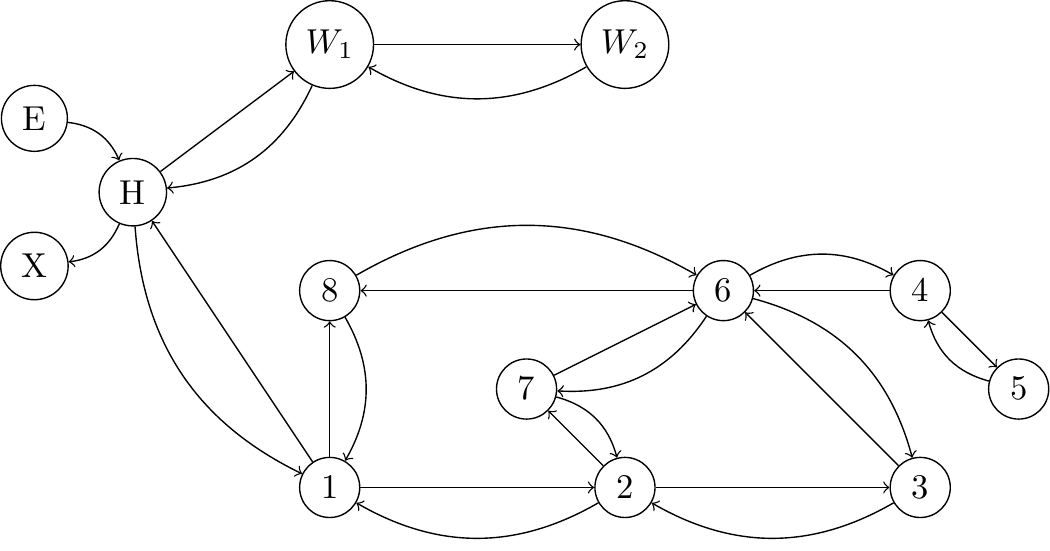}
\caption{A possible graph for the Musée de l'Orangerie}
\label{orangerie-graph}
\end{figure}

The Musée de l'Orangerie can be represented as the graph presented in Figure \ref{orangerie-graph}. We can see that there is only one entrance and one exit in the museum and they are located at the same place. Therefore, we consider the entrance and the exit as two different vertices in the graph to facilitate the model. The mapping between vertices and rooms is shown in Table \ref{orangerie-vertex-to-room}.

\begin{table}[tbp]
\caption{The Musée de l'Orangerie: vertices and rooms}
\label{orangerie-vertex-to-room}
\centering
\begin{tabular}{|c|l|}
\hline
Vertex & Room \\ \hline
E & Entrance \\ \hline
X & Exit \\ \hline
H & Hall \\ \hline
$W_1$ & Water Lilies (first part) \\ \hline
$W_2$ & Water Lilies (second part) \\ \hline
1 & L'Age d'Or \\ \hline
2 & Paul Guillaume Room \\ \hline
3 & Impressionism \\ \hline
4 & Modern primitives \\ \hline
5 & Laurencin room \\ \hline
6 & Modern classicism \\ \hline
7 & Derain room \\ \hline
8 & Soutine / Utrillo room \\ \hline
\end{tabular}
\end{table}

\subsection{Personalized tour problem modeling}

For the sake of satisfying the visitor maximally, a visit tour should be proposed according to the visitor’s preferences and constraints.

A personalized tour problem can be defined as a 6-tuple $\langle G, I, R, u, t, T_{MAX} \rangle$ where:
\begin{itemize}
\item $G$ is the museum graph representing the museum topology as defined above.
\item $I$ is the set of artworks which have to be included in the tour.
\item $R$ is the set of artworks which have to be excluded of the tour.    
\item $u$ is a mapping $u: P \rightarrow \mathbb{R^+}$. For each artwork $p \in P$, $u(p)$ denotes the interest of the visitor for the artwork $p$.
\item $t$ is a mapping $ t: A \cup V \cup P \rightarrow \mathbb{R^+} $. For each room, arc and artwork, we have a time to spend. It can be the time needed to cross a room or an arc. It can also be the time to see an artwork.
\item $T_{MAX}$ is the maximum time that a visitor wants to spend in the museum.
\end{itemize}

A visit tour may be a simple path without any cycles (an elementary path) or a sophisticate path including cycles (a non-elementary path).

\newpage

We define a tour as a sequence of pairs $\langle v, P_v \rangle$ where $v \in V$ and $P_v \subseteq r(v)$ (note that $P_v$ may be $\emptyset$ because we can cross a room without seeing any artwork).
A tour $ T = \left( \langle v_1, P_{v_i} \rangle, ..., \langle v_n, P_{v_n} \rangle \right) $ is a solution to the personalized tour recommendation problem when:

\begin{enumerate}
\item The vertex $v$ in the first element of $T$ is an element of $E$ (the tour starts by an entrance).
\item The vertex $v$ in the last element of $T$ is an element of $X$ (the tour ends by an exit).
\item All consecutive elements $T_1$ and $T_2$ of $T$ share the same vertex or an arc $a_{ij} \in A$ must exist from the vertex $i$ of $T_1$ to the vertex $j$ of $T_2$.
\item The total time required to see all the artworks and pass through all the rooms (and ways) is not bigger than $T_{MAX}$.
\end{enumerate}

\subsubsection*{Application to the Musée de l'Orangerie}

we may have a visit tour like the following:
$$\left( \langle E, \emptyset \rangle, \langle H, \emptyset \rangle, \langle W_1, p_1 \rangle, \langle W_2, p_5 \rangle, \langle W_1, \emptyset \rangle, \langle H, \emptyset \rangle, \right. $$
$$ \left. \langle 1, p_{10} \rangle, \langle 8, p_{103} \rangle, \langle 1, \emptyset \rangle, \langle H, \emptyset \rangle, \langle X, \emptyset \rangle \right)$$
In this tour, the visitor should cross the receiving hall $H$ three times, exhibition room $W_1$ and $1$ twice respectively. Although we may traverse a room several times, the visitor is supposed to visit the room only once. Consider for instance the exhibition room $W_1$, we may visit the selected artworks when we reach this room for the first time. The second time, we would just cross the room to visit another one.


\section{Integer Linear Programming approach to solve the Optimal Personalized Tour Recommendation Problem}
\label{sec:model}

Before introducing an ILP model to solve the Personalized Tour Recommendation Problem, we define several decision variables:

\begin{itemize}
\item $x_p $ equals $1$ if the artwork $p \in P$ is included in the proposed tour, $0$ otherwise.
\item $y_a $ equals $1$ if arc $a \in A$ is crossed in the proposed tour, $0$ otherwise.
\item $f_a $ denotes the number of rooms crossed when we arrive at arc $a \in A$ in the visit walk.
\item $z_v $ equals $1$ if room $v \in V$ is traversed in the proposed tour (no matter whether we visit an artwork of this room or not), $0$ otherwise.
\end{itemize}

\vfill
\pagebreak

Given a personalized tour problem (as defined in section \ref{sec:personalized-tour-recommendation-problem}), the objective function of the Optimal Personalized Tour Recommendation Problem (OPTRP) is to maximize the overall satisfaction of the proposed visit tour for the visitor:


\begin{equation}
\max  \sum_{p \in P} x_p \times u(p) \quad \quad \textrm{OPTRP}
\end{equation}
s.t.
\begin{align}
\label{eqn: entrance}
& \sum_{v \in E} \sum_{a \in \delta^+(v)} y_a  =  1 &&\\
\label{eqn: exit}
& \sum_{v \in X} \sum_{a \in \delta^-(v)} y_a = 1  &&\\
\label{eqn: room}
& \sum_{a \in \delta^+(v)} y_a = \sum_{a \in \delta^-(v)} y_a, && \forall v \in V \setminus (E \cup X)\\
\label{eqn: flow-room-relation1}
& f_a  \geq  y_a, &&  \forall a \in A \\
\label{eqn: flow-room-relation2}
& f_a  \leq |V| \times y_a, && \forall a \in A\\
\label{eqn: flow-room-relation3}
& \sum_{a \in \delta^-(v)} f_a  =  \sum_{a \in \delta^+(v)} f_a - z_v, && \forall v \in V \setminus (E \cup X)\\
\label{eqn: arc-room-relation1}
& \sum_{a \in \delta^+(v)} y_a + \sum_{a \in \delta^-(v)} y_a \geq z_v , && \forall v \in V\\
\label{eqn: arc-room-relation2}
& y_a  \leq z_v, && \forall v \in V, \forall a \in \delta^+(v) \cup \delta^-(v)\\
\label{eqn: artwork-room-relation}
& x_p \leq z_v,  && \forall v \in V, \forall p \in \left\lbrace p : r(p) = v\right\rbrace \\
\label{eqn: artwork-required}
& x_p = 1, && \forall p \in I
\\
\label{eqn: artwork-excluded}
& x_p = 0, && \forall p \in R
\end{align}
\begin{equation}
\label{eqn: time}
\sum_{v \in V} z_v \times t_v + \sum_{a \in A} y_a \times t_a  + \sum_{p \in P} x_p \times t_p \leq T_{MAX}
\end{equation}

Constraints (\ref{eqn: entrance}) and (\ref{eqn: exit}) ensure that the visitor should enter a museum from a unique entrance and finish the visit by a unique exit respectively (this model considers the case of multiple entrances and exits). Constraint (\ref{eqn: room}) makes sure that a visitor should exit a room $v$ after crossing or visiting it.  Constraint (\ref{eqn: flow-room-relation1}) expresses that a visitor should have crossed at least a room before arriving at an arc $a$, while constraint (\ref{eqn: flow-room-relation2}) imposes that no flow is moving on the arc $a$, if it is not crossed in the visit tour. Constraint (\ref{eqn: flow-room-relation3}) means that if a room $v$ is crossed in the tour, then the number of rooms crossed before arriving at this room equals the number of rooms crossed after leaving $v$ minus one. Otherwise, they should be equal, since the room will not appear in the tour. Constraint (\ref{eqn: arc-room-relation2}) imposes that a room $v$ should be crossed as long as one input arc or one outgoing arc is crossed.  Constraint (\ref{eqn: arc-room-relation1}) ensures that a room $v$ should not be crossed if none of the input arc or output link is used during the visit. Constraint (\ref{eqn: artwork-room-relation}) indicates that if a room $v$ is not crossed, none of its artworks will be proposed for visiting. Constraints (\ref{eqn: artwork-required}) and (\ref{eqn: artwork-excluded}) ensure that an artwork should be included or excluded from the proposed tour if the visitor asks for it. The last constraint (\ref{eqn: time}) guarantees that the time spent in front of the artworks and the time required to pass through rooms (and ways) does not exceed the available time for the visitor.

The ILP model we propose provides a visit tour starting from an entrance and terminating at an exit. In \cite{artois_2012}, authors also proposed an ILP model to plan the personalized visit. They divided the studied proposed into two sub-problems: first determine the number of moves (denoted as $K$) for a complete walk in the museum graph, and then solve the museum routing problem while maximizing visitor satisfaction. Since both of these sub-problems are NP-Hard, authors of \cite{artois_2012} proposed to solve both of them by constraint programming. The complexity of their model depends a lot on $K$, which is generally large (at least equals to $|V|$). To compare the complexity of our model with the ILP mode in \cite{artois_2012}, the number of variables and constraints are listed in Table \ref{table_comparison}.

\begin{table*}[tbp]
\centering
  \caption{Comparison of ILP Models}
\label{table_comparison}
\begin{tabular}{|c||c|c|c|}
\hline
Terms    & OPTRP ILP & ILP \cite{artois_2012} \\
\hline
\hline
Variables             & $x_p$, $y_a$, $f_a$, $z_v$             & $x_p$, $x_{p,k}$, $c_{i,j,k}$\\
Number of variables   & $|P|+2|A| + |V|$          &  $(|P|+|A|) \times K + |P|$ \\
Constraints           & (\ref{eqn: entrance})-(\ref{eqn: time})    &   (2)-(8) in \cite{artois_2012}\\
Number of constraints &    $3+4|A|+|P| + 3|V|+ |I|- 2(|E|+|X|)$  & $|A|\times(K-1)+ |P|\times(2K-1) + K + 3$\\
\hline
\end{tabular}
\end{table*}

\section{Visitor preferences modeling}
\label{sec:visitor-preferences-modeling}

The interest function $u$ should reflect the satisfaction of the visitor for each artwork $p \in P$. The nearer to his preferences is an artwork $p$, the greater is $u(p)$.

\subsection*{Representation of artworks and the visitor preferences}

We define $C_p$ as the set of all characteristics of an artwork $p$ and $C$ as the union of all these sets (i.e. $ C = \left\lbrace c | c \in C_p \forall p \in P \right\rbrace $). A characteristic may be the theme, the type of support, the date of creation, the name of the artist or anything that identify an artwork.

We can represent any artwork $p \in P$ as a caracteristics vector $v_p = (v_{p_1}, ..., v_{p_n})$ in a vector-space of $|C|$ dimensions. Each element $v_{p_i} \in \mathbb{R}^+$ in the vector is a numerical value measuring the relevance of the artwork $p$ to the associated characteristic.
Additionally we define a vector $v$ in the same vector-space as the vector representing the visitor preferences (where each element of $v$ measures the interest of the visitor for the associated characteristic).

\subsection*{Measuring the interest for an artwork}

To identify the interest for the visitor to an artwork, we compare $v$ and $v_p$ with the cosine similarity which calculate the angle between two vectors. The formula is the following:

\begin{equation}
similarity(v_p, v) = \frac{\sum_i^n v_{p_i} \times v_i}{\sqrt{\sum_i^n v_{p_i}^2} \times \sqrt{\vphantom{\sum_i^n v_{p_i}^2} \sum_i^n v_i^2}}
\end{equation}

The resulting similarity ranges from $0$, meaning that the visitor is not interested at all by the artwork, to $1$ meaning that the artwork exactly matches his preferences.

In our model, we used $u(p) = similarity(v_p, v)$ where $v_p$ and $v$ are the vectors of the artwork $p$ and the visitor preferences respectively.

\section{Simulations and numerical results}
\label{sec:simulations-numerical-results}

We implemented the ILP model described in section \ref{sec:model} by using the IBM CPLEX 12 library\footnote{http://www-01.ibm.com/software/commerce/optimization/cplex-optimizer/}.

The program takes several input parameters:

\begin{itemize}

\item The graph modeling the museum as defined in section \ref{sec:personalized-tour-recommendation-problem}

\item The interest function $f$ to use. This function produces interest vectors as defined in section \ref{sec:visitor-preferences-modeling}

\item The maximum duration that a visitor can spend in the museum

\end{itemize}

It outputs the proposed tour (as defined in section \ref{sec:personalized-tour-recommendation-problem}).

\subsection{Intrinsic interest}

As we saw in section \ref{sec:enertex}, the Enertex algorithm ranks the sentences of a document. We used Enertex as the following:

\begin{enumerate}

\item From the website of the museum, we created an XML file containing the following information for each artwork: the title, the artist name, the year and the description of the artwork.

\item We extracted data from the XML to produce a file where each pseudo sentence is a concatenation of title, artist and description.

\item The latter file is used as an input to Enertex with the query "musée orangerie peinture impressionniste postimpressionniste" to drive the balancing process of the system.

\end{enumerate}

It produces a ranking for artworks depending on the information displayed by the museum (for each artwork, the result is a value ranging from 0 to 1).

\begin{figure}[h]
\centering
\includegraphics[scale=0.55]{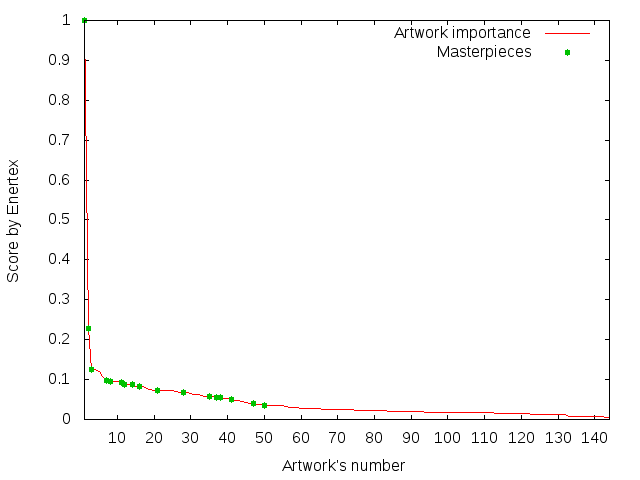}
\caption{Ranking by Enertex of the artworks in the Musée de l'Orangerie}
\label{fig:orangerie-ranking}
\end{figure}

Fig. \ref{fig:orangerie-ranking} shows the ranking of the artworks in the \textit{Musée de l'Orangerie} provided by Enertex.

As we can see, the resulting ranking is in agreement with the information provided by the museum. Indeed the masterpieces (according to the website of the museum) represent the most important artworks (which have the highest scores).

\newpage

\subsection{Interest functions}

Four different interest function were designed to simulate the visitor preferences.

\begin{itemize}

\item $f_1$: produces the same vector $v = (1)$ for each artwork and visitor preferences

\item $f_2$: produces a vector $v_p = (s_p)$ where $s_p$ is the score given by Enertex for the artwork $p$ and produces a vector $v = (1)$ as the visitor preferences.

\item $f_3$: produces vectors $v = (v_1, ..., v_n)$ of size $n$ equals to the number of artists. Each artwork is represented as a vector where $v_i = 1$ if the artwork is created by the artist $i$, $0$ otherwise. The visitor preferences are represented as a vector where $v_i = 1$ if the visitor is interested by the artist $i$, $0$ otherwise.

\item $f_4$: produces vectors $v = v_{f_2} \| 10 \times v_{f_3}$ where $v_{f_2}$ and $v_{f_3}$ are the vectors produces by $f_2$ and $f_3$ respectively.

\end{itemize}

The first function defines the baseline: the visitor has no interest at all. The second uses the ranking provided by Enertex: the visitor wants to discover the most important artworks of the museum. The third uses the visitor preferences (a set of interesting artists). The last combines both visitor preferences and museum point of view (we multiply by $10$ because we want to give more importance to the visitor preferences than to the museum point of view).

\subsection{Evaluation}

To evaluate the output tour, we measure the relevance percentage defined as :

\begin{equation}
rp = \frac{100 * \sum_{p \in T}{f(p)}}{\sum_{p \in P}{f(p)}}
\end{equation}

where $T$ is the set of artworks proposed in the tour (i.e. $T \subseteq P$) and $f$ the interest function used (as saw above). The relevance percentage $rp$ denotes a satisfaction rate of the visitor.

\subsection{Results}

For each function $f$, we ran the program with different time limits from 30 to 330 minutes (the time required to visit the entire collection) by steps of 15 minutes. For $f_2$ and $f_3$, we randomly pre-generated 5,000 combinations of 2, 3, 4 and 5 artists (i.e. the same combinations are used with $f_2$ and $f_3$) and calculated the arithmetic mean relevance percentage for each duration.

\begin{figure*}[!htb]
\centering
\includegraphics[scale=1]{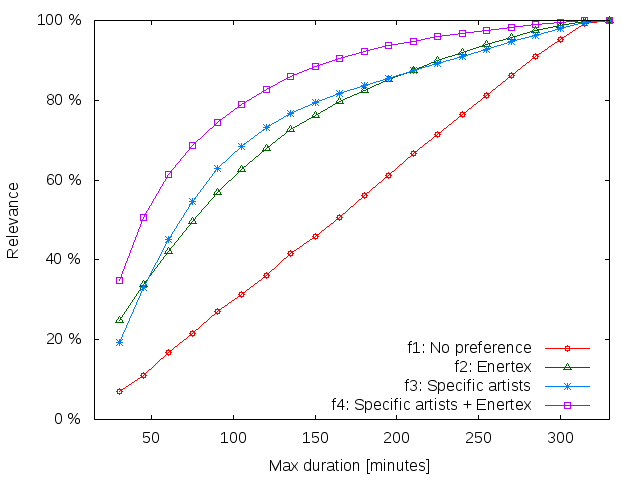}
\caption{Evolution of the relevance percentage}
\label{orangerie-results}
\end{figure*}

Figure \ref{orangerie-results} shows the evolution of $rp$ for each function $f$. As expected, the first interest function $f_1$ produces a linear evolution of the relevance percentage (given that all artworks have the same interest, the tour includes the greatest number of artworks). With Enertex we are able to propose efficent tours to visitors who want to discover the museum (without particular preferences). The combination of both visitor preferences and intrinsic preferences produces the best results up to 49\% of relevance improvement. It also appears that after 150 minutes, the improvement is less significant, we could assume that, from the visitor point of view, the optimal tour duration is about 2 hours and a half.

\section{Conclusion and future works}
\label{sec:conclusion}

This research tackles the problem of optimizing museum visits according to visitors preference and artwork importance. As a first milestone for next works taking into account the individual behavior in museum visits, it sets an original model combining computational optimisation and automatic learning via artificial intelligence. We first drew the optimization framework based on graph theory to depict the spatial organization of the museum (including rooms and paths), that requires an Integer Linear Programming to maximize the visitor overall satisfaction and to generate an optimal path, that is to say a series of rooms and artworks to be seen by the visitor. In complement, we compute an artwork description analysis by a natural language processing based on textual energy (using an algorithm called Enertex). This leads to ranking the different artworks according to the descriptions given by the museum, related to their artistic importance. 
Associating those two complementary approaches, we are then able to design optimal paths for visitors according to different interest functions based on artwork objective values assigned by museums.

Future works concern more subjective behavior of visitors depending on their profiles and leisure practices. Indeed, the project aims at finding relevant recommendations for optimal visit tours that rise a better fitness between the visitor wishes and the museum artistic supply. We can think about using natural language processing to generate the set of characteristics for all the artworks in a museum and calculate better interest vectors but also produce a summary of the proposed tour.

This information may advantageously be used by existing and potential visitors to refine the way they get involved in their cultural pratices. Indeed, it is admitted that the museum connoiseurs use to develop a critical mind about new services in a numerical society. Thence, aware visitors become able to appreciate the personalized routing recommendation system provided by their prefered museums.

\FloatBarrier
\section*{Acknowledgments}

The authors would like to thank the \textit{Departement du Vaucluse} (France) and the \textit{FR Agorantic} for the financial supports (projects \textit{@MUSE} and \textit{InfoMuse}).

\bibliographystyle{IEEEtran}
\bibliography{IEEEabrv,references}

\end{document}